\documentclass[10pt,twocolumn,letterpaper]{article}

\usepackage[pagenumbers]{cvpr} 
\usepackage[accsupp]{axessibility}  

\usepackage{amsmath}
\usepackage{amssymb}

\usepackage[pagebackref,breaklinks,colorlinks]{hyperref}

\usepackage[capitalize]{cleveref}
\crefname{section}{Sec.}{Secs.}
\Crefname{section}{Section}{Sections}
\Crefname{table}{Table}{Tables}
\crefname{table}{Tab.}{Tabs.}

\usepackage[utf8]{inputenc} 
\usepackage[T1]{fontenc}    
\usepackage{hyperref}       
\usepackage{url}            
\usepackage{booktabs}       
\usepackage{amsfonts}       
\usepackage{nicefrac}       
\usepackage{microtype}      
\usepackage{xcolor}         
\usepackage{array}
\usepackage{graphicx}
\usepackage{mathrsfs}
\usepackage{pifont}
\usepackage{comment}
\usepackage{tabulary,multirow,xspace}
\usepackage{fixmath,mathtools,mmstyle}
\usepackage{cite}
\usepackage{subcaption}
\usepackage{caption}
\usepackage{float}
\newcommand{\myparagraph}[1]{{\noindent\bf #1}}

\def\cvprPaperID{6770} 
\def\confName{CVPR}
\def\confYear{2022}

\begin{document}

\title{Exploring Set Similarity for Dense Self-supervised Representation Learning}

\author{
  Zhaoqing Wang$^{1}$\quad
  Qiang Li$^{2,}$\thanks{Corresponding author.}\quad
  Guoxin Zhang$^2$\quad
  Pengfei Wan$^2$ \\
  Wen Zheng$^2$\quad
  Nannan Wang$^3$\quad
  Mingming Gong$^4$\quad
  Tongliang Liu$^1$ \\
  \small $^1$University of Sydney\quad
  \small $^2$Kuaishou Technology\quad
  \small $^3$Xidian University\quad
  \small $^4$University of Melbourne\\
  \tt\small \{derrickwang005,leetsiang.cloud\}@gmail.com\\
  \tt\small \{wanpengfei,zhengwen\}@kuaishou.com;
  \tt\small zgx.net@qq.com;
  \tt\small nnwang@xidian.edu.cn\\
  \tt\small mingming.gong@unimelb.edu.au;
  \tt\small tongliang.liu@sydney.edu.au
  }
\maketitle

\begin{abstract}
    By considering the spatial correspondence, dense self-supervised representation learning has achieved superior performance on various dense prediction tasks. 
    However, the pixel-level correspondence tends to be noisy because of many similar misleading pixels, e.g., backgrounds.
    To address this issue, in this paper, we propose to explore \textbf{set} \textbf{sim}ilarity (\textbf{SetSim}) for dense self-supervised representation learning.
    We generalize pixel-wise similarity learning to set-wise one to improve the robustness because sets contain more semantic and structure information.
    Specifically, by resorting to attentional features of views, we establish the corresponding set, thus filtering out noisy backgrounds that may cause incorrect correspondences.
    Meanwhile, these attentional features can keep the coherence of the same image across different views to alleviate semantic inconsistency.
    We further search the cross-view nearest neighbours of sets and employ the structured neighbourhood information to enhance the robustness.
    Empirical evaluations demonstrate that SetSim surpasses or is on par with state-of-the-art methods on object detection, keypoint detection, instance segmentation, and semantic segmentation.
\end{abstract}

\begin{figure}[t]
	\centering
    \includegraphics[width=1.\linewidth]{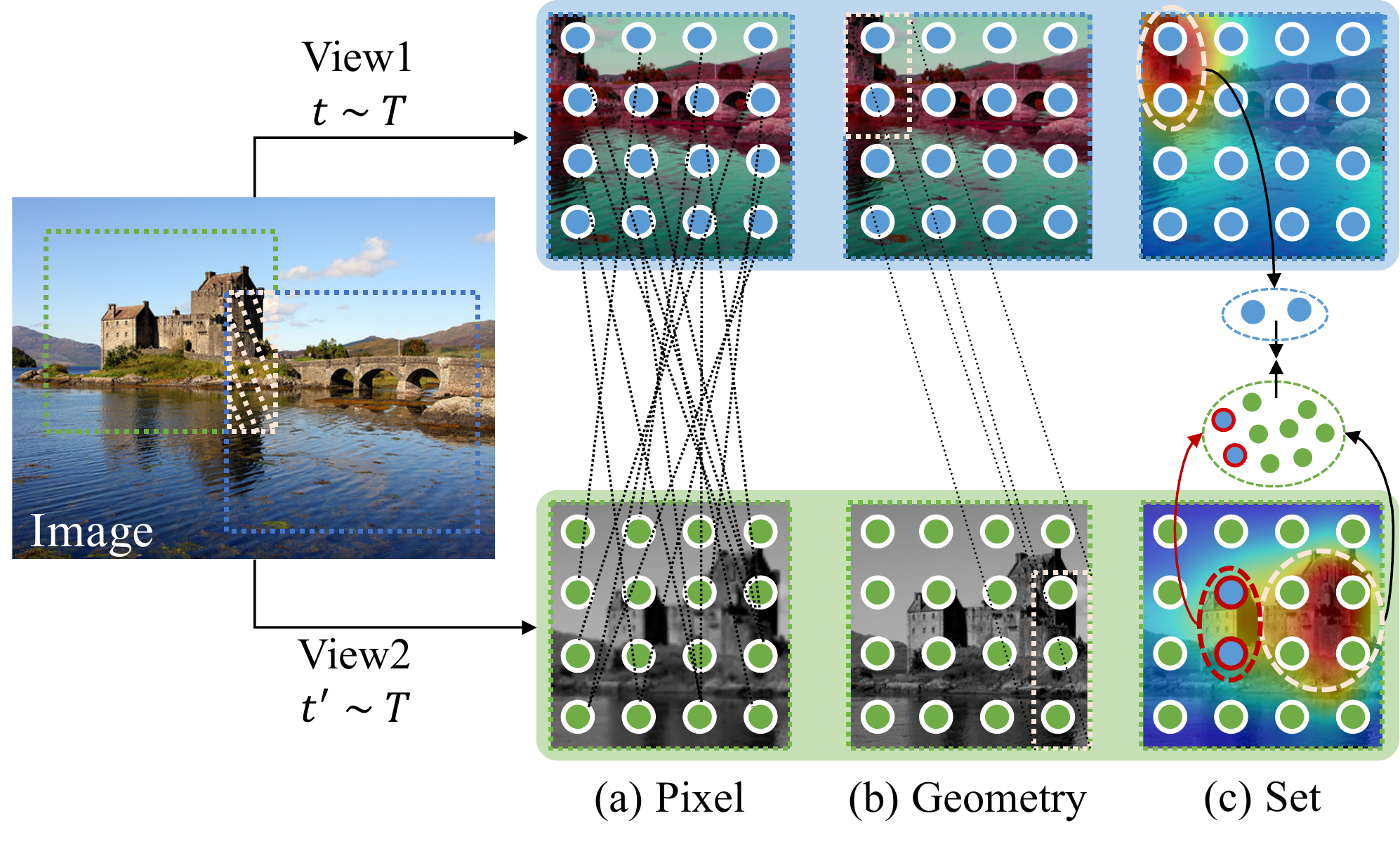}
	\caption{\textbf{The comparison of existing pixel-wise correspondence with our proposed method.}
            (a) Pixel-based: compare all combinations of pixel-wise features and maximize the most similar pairs.
            (b) Geometry-based: force features in overlapping regions to remain constant and distinguish features from different locations.
            (c) Set-based: considering misleading features and semantic inconsistency in the former methods,
            we propose to explore set similarity across two views for dense self-supervised representation learning.
            By resorting to attentional features, SetSim constructs corresponding sets across two views (\textcolor{blue}{blue point} \& \textcolor{green}{green point}),
            which can filter out misleading features and keep the coherence of the same image across views.
            Furthermore, SetSim searches the cross-view nearest neighbours (\textcolor{blue}{blue point} with \textcolor{red}{red circle}) to enhance the structured neighbourhood information.}
    \label{fig:Introduction}
	\vspace{-2.0mm}
\end{figure}

\section{Introduction}\label{sec:Introduction}

Pretraining has become a widely-used paradigm in various computer vision tasks. Generally, models are first pretrained on large-scale datasets (e.g., ImageNet \cite{deng2009imagenet}) and fine-tuned on the specific tasks.
Recently, self-supervised pretraining has broken the dominance of the supervised ImageNet \cite{deng2009imagenet} pretraining on almost all downstream tasks including image classification \cite{he2016deep}, object detection\cite{ren2015faster}, semantic segmentation \cite{fu2019dual,fu2020scene}, etc.
In particular, state-of-the-art self-supervised representation learning methods \cite{caron2018deep,caron2020unsupervised,chen2020exploring,grill2020bootstrap} mainly adopt the \textit{instance discrimination} formulation as their pretext task to obtain transfer learning ability for downstream tasks.
The main idea of these methods is to maximize the similarity of two data-augmented views of the same image, while minimizing the similarity of views generated from different images.

Since most of the self-supervised methods are designed for image-level tasks like image classification, they are sub-optimal for dense prediction tasks, e.g., object detection and semantic segmentation. To narrow this performance gap, dense self-supervised learning has been explored recently \cite{wang2020DenseCL,xie2020propagate,pinheiro2020unsupervised}.
A popular way is to leverage intrinsic spatial information that a matching pair of pixel-wise features should remain constant over different viewing conditions.
DenseCL \cite{wang2020DenseCL} compares all combinations of pixel-wise features across two views and picks out the most similar pairs as spatial correspondences.
Besides, VADeR \cite{pinheiro2020unsupervised} and PixPro \cite{xie2020propagate} adopt geometry-based correspondence, which means that two views' pixel-wise features from the same location of the same image are treated as positive pairs, while features obtained from different locations are treated as negative pairs.

Nonetheless, in general, the pixel-wise correspondence based on similarity or geometry information is more likely to be noisy.
Specifically, if there is a pixel-wise feature vector, there would be many misleading features similar to it, thus causing incorrect correspondences, which are illustrated in Figure~\ref{fig:Introduction}(a).
Besides, in Figure~\ref{fig:Introduction}(b), the overlapping region of two views splits similar semantic features into two parts, which are pushed far in the embedding space by a learning objective, resulting in spatial semantic inconsistency.
Although recently proposed methods have shown promising transfer performance improvements on dense prediction tasks,
the issue of establishing robust correspondence remains unsolved, therefore, we are motivated to seek further exploration on this issue in order to apply it to our research.

In this paper, we propose to explore \textbf{set} \textbf{sim}ilarity (SetSim) across views for dense self-supervised representation learning.
Considering that a set of pixel-wise features can represent more semantic and structure information than individual counterpart, we generalize pixel-wise similarity learning to set-wise one to improve the robustness.
In particular, based on the attention map, we construct the corresponding set, which contains pixels of two different views with similar semantic information.
As shown in Figure~\ref{fig:Introduction}(c), attention maps are able to reveal the salient objects, \ie, castle, in two views of the input image, which effectively keep the coherence of the input image.
Since certain useful pixels are excluded from the set, we further search the cross-view nearest neighbours of one view's set and enhance the structured neighbourhood information.
Finally, the model maps the pixels in the corresponding set to similar representations in the embedding space by a contrastive \cite{he2020momentum, li2022selective} or cosine similarity \cite{chen2020exploring} optimization function.

The proposed SetSim outperforms or is on par with the state-of-the-art methods on various dense prediction tasks, including object detection, keypoint detection, instance segmentation, and semantic segmentation.
Compared to the MoCo-v2 baseline, our method can significantly improve the localization and classification abilities on dense prediction tasks:
$+2.1 AP^{b}$ on VOC object detection,
$+1.3AP^{b}$ on COCO object detection, $+0.4AP^{kp}$ on COCO keypoint detection,
$+1.0AP^{m},\, +2.2AP^{m}$ on COCO and Cityscapes instance segmentation, 
$+2.6mIoU$ on VOC object segmentation, $+1.7mIoU,\, +1.6mIoU$ on ADE20K and Cityscapes semantic segmentation, respectively.

\section{Related Work}

\myparagraph{Self-supervised representation learning.}
Self-supervised representation learning is a kind of unsupervised representation learning, which has received extensive attention in recent years.
It leverages the intrinsic structure of data as a supervisory signal for training and learns informative and transferable representations for downstream tasks.
Early self-supervised learning approaches consist of a wide range of pretext tasks, including denoising \cite{vincent2008extracting}, inpainting \cite{pathak2016context}, colorization \cite{zhang2016colorful, larsson2017colorization}, egomotion prediction \cite{agrawal2015learning}, and so on \cite{mundhenk2018improvements,donahue2016adversarial,zhang2017split}.
Besides, a series of high-level pretext tasks are under research, such as rotation \cite{gidaris2018unsupervised}, jigsaw puzzles \cite{noroozi2016unsupervised}, predicting context \cite{doersch2015unsupervised} and temporal ordering \cite{misra2016shuffle}.
However, these methods achieved minimal success in computer vision.

Recently, contrastive self-supervised learning has emerged as a promising approach to unsupervised visual representation learning.
The breakthrough one is SimCLR \cite{chen2020simple}, which adopts the \textit{instance discrimination} formulation as its pretext task.
It generates two views of each image by a diverse set of data augmentations and maximize the similarity of two augmented views from the same image, meanwhile, minimizing similarities with a large set of views from other images.
Besides, MoCo \cite{he2020momentum,chen2020improved} introduces a momentum encoder to improve the consistency of a queue of negative samples and achieves remarkable performance.
Shortly after that, another category of methods based on clustering \cite{caron2020unsupervised, caron2018deep,caron2019unsupervised,asano2019self} is proposed, which alternates between clustering feature representations and learning to predict the cluster assignment.
More recently, BYOL \cite{grill2020bootstrap} and SimSiam \cite{chen2020exploring} are came up with directly predicting the output of one view from another view without consideration of negative samples.
Nonetheless, image-level self-supervised pretraining can be sub-optimal for dense prediction tasks due to the discrepancy between image-level and pixel-level prediction. 

\myparagraph{Dense self-supervised representation learning.}
Image-level supervised and self-supervised pretraining has achieved encouraging results on a series of downstream tasks, including image classification, object detection, semantic segmentation and so on.
Nonetheless, previous studies \cite{sun2019deep,he2019rethinking,tan2020efficientdet} demonstrate that there is a transfer gap between image-level pretraining and dense prediction tasks.
Recently, several related approaches \cite{wang2020DenseCL,xie2020propagate,pinheiro2020unsupervised} are proposed to explore dense self-supervised representation learning.
They generalize the instance discrimination from image-level to pixel-level.
To be specific, the positive pairs of local features across views are defined by pixel-level correspondences and features excluded from the correspondence are treated as the negative pairs.
In particular, DenseCL \cite{wang2020DenseCL} compares all combinations of feature vectors and pulls the most similar pairs closer, which is similar to clustering.
By utilizing parameters of the affine transformation, VADeR \cite{pinheiro2020unsupervised} and PixPro \cite{xie2020propagate} map corresponding pixel-wise features in each view to their associated features.
However, pixel-wise correspondence is likely to be noisy because there is a wide range of highly misleading pixel-wise features \cite{xia2021robust,xia2020part,xia2019anchor,xia2021sample}, while the geometric correspondence is difficult to capture the coherence between two views of the input image.
In this work, we introduce set similarity for dense self-supervised representation learning to improve the robustness.


\begin{figure*}[th]
    \begin{center}
        \includegraphics[scale=0.52]{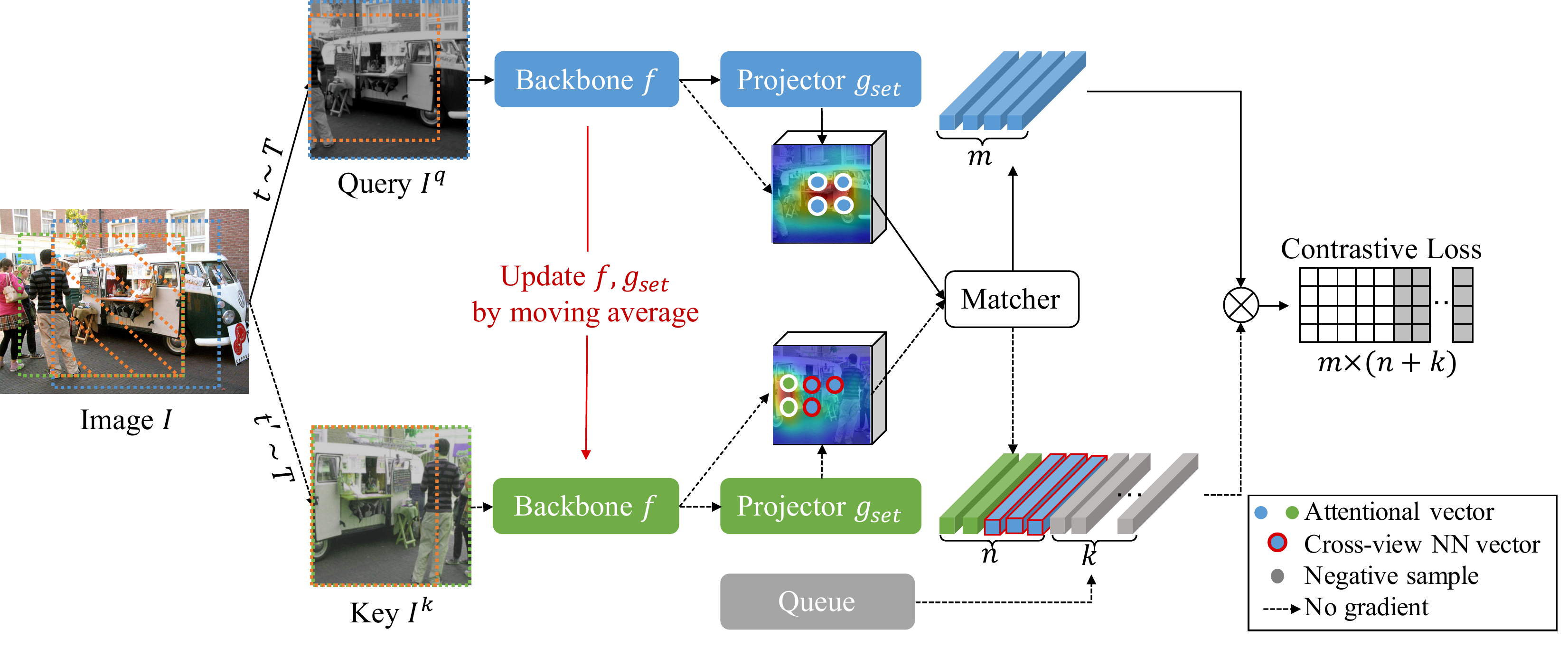}
        \caption{\textbf{An overview of the architecture of the proposed SetSim.}
        The SetSim architecture has two branches, including an encoder and its momentum-updated one.
        SetSim takes two augmented views of an image as inputs, $I^{q}$ and $I^{k}$. 
        For each view, the backbone $f$ is adopted to extract visual features, and a convolutional projector $g_{set}$ is adopted to generate transformed feature maps.
        SetSim introduces a matcher to construct corresponding sets and define the correspondence between two views.
        Finally, SetSim is optimized by an image-level and a set-level contrastive loss in an end-to-end manner.
        For brevity, the image-level branch \cite{he2020momentum} is not shown in the figure.}
    \label{fig:method}
    \end{center}
\end{figure*}

\section{Methodology}\label{sec:methods}
In this section, we first revisit the instance discrimination task and its general pipeline for self-supervised representation learning.
Subsequently, we present a detailed description of our proposed SetSim, \ie, a dense self-supervised learning framework based on set similarity.

\subsection{Preliminaries}

Instance discrimination is a widely-used pretext task for self-supervised visual representation learning \cite{chen2020simple,he2020momentum,wu2018unsupervised,chen2020improved}.
Given an unlabeled dataset, the input image $I$ is augmented by a series of pre-defined data augmentations $T=[T_{1},T_{2}, ..., T_{n}]$. By sampling $t \sim T$ and $t^\prime \sim T$, we can generate the query view $I^{q} = t(I)$ and key view $I^{k} = t^\prime(I)$.
For each view, an encoder is adopted to extract image-level features $p_{img}$.
The encoder consists of two main components, the backbone $f$ and the projector $g$.
Notice that only the backbone is transferred to downstream tasks after the pretraining process.
Subsequently, a contrastive loss is adopted to pull each encoded query $p_{img}^{q}$ close to its positive encoded key $p_{img}^{k_{+}}$ and away from its negative encoded keys $p_{img}^{k_{-}}$:
\begin{equation}
    \mathcal{L}_{img} = -\log \frac{\exp{(p_{img}^{q} \cdot p_{img}^{k_{+}} / \tau)}}{\sum_{p_{img}^{k}} \exp(p_{img}^{q} \cdot p_{img}^{k} / \tau)},
    \label{eq1}
\end{equation}
where encoded features $p_{img}$ are L2-normalized, and $\tau$ is a temperature hyper-parameter \cite{wu2018unsupervised}.

\subsection{Architecture Overview}
As illustrated in figure~\ref{fig:method}, the proposed SetSim framework mainly consists of four parts: a backbone, two projectors, a matcher, and a queue.
Firstly, SetSim augments an input image into two data-augmented views, $I^{q}$ and $I^{k}$.
Then, each view's deep feature are generated by the backbone network $f$ (a ResNet-50 \cite{he2016deep} is used by default), and are fed into two parallel projectors $g_{img}$ and $g_{set}$.
To maintain the basic architecture, we keep the design of the set-level projector as simple as the image-level one.
Specifically, the image-level projector $g_{img}$ is composed of two fully connected layers with a ReLU layer between them, and the set-level projector $g_{set}$ is composed of two $1 \times 1$ convolution layers with a ReLU layer between them.
Upon the attended transformed features, SetSim employs a matcher to establish two corresponding sets spatially across two views.
Finally, we adopt a standard contrastive loss function \cite{he2020momentum} for image-level optimization and a modified contrastive loss function for set-level optimization.

\subsection{Set Similarity Dense Representation Learning}

\myparagraph{Constructing Corresponding Set.}
With the help of image-level contrastive loss, attention maps of the top layers can reflect some salient regions (e.g., objects or stuff), which is crucial to alleviate both effects of misleading pixel-wise features and semantic inconsistency.
In particular, for each data-augmented view from the same input image $I$,
the backbone $f$ extracts feature maps $z \in \mathbb{R}^{C \times HW}$ and the convolutional projector $g_{set}$ generates feature maps $p \in \mathbb{R}^{C^\prime \times HW}$,
which is formulated as:
\begin{equation}
    p = g_{set}(z), \quad z = f(I),
    \label{eq2}
\end{equation}
where the feature maps before and after projection have different channel dimension $C$ and $C^\prime$, respectively.
To construct the corresponding set, we first obtain the spatial attention map $A$ by computing statistics of feature maps $z$ across the channel dimension $C$,
which is formulated as:
\begin{equation}
    A = \sum_{i=1}^{C}|z_{i}|,
\end{equation}
where $z_{i} = z(i,:)$ and the absolute value operation is element-wise.
Then, we employ a relative selection strategy to append attentional vectors into the corresponding set, which uses Min-Max normalization for rescaling $A$ and introduces a threshold $\delta$ for selecting vector $p_j$,
which is defined as:
\begin{equation}
    A^\prime = \frac{A - \min(A)}{\max(A) - \min(A)} ,
    \label{eq3}
\end{equation}
\begin{equation}
    \Omega = \{j: A^\prime(j) \geq \delta\},
    \label{eq4}
\end{equation}
where $A^\prime$ is the resacled attention map, $j$ is the spatial index of feature maps $p$ thus $p_{j} \in \mathbb{R}^{C^\prime}$.
We conduct the discussion of $\delta$ in the following experiment section.
Finally, attentional feature vectors $p_{j}$ can be adaptively appended in the  set $\Omega$.

\myparagraph{Set2Set-NN Matching Strategy.}
By resorting to attention maps, SetSim generates the corresponding set of attentional feature vectors from two views.
For simple illustration, we assume that the numbers of attentional vectors of the query and key view are $m$ and $n$, respectively.
For each attentional query vector $p_{i}^{q}$, we first establish fully-connected correspondences $s_{i}$ with each attentional key vectors $p_{i}^{k}$.

Due to the threshold $\delta$ selection, some useful vectors could be excluded from the corresponding set.
So, we further search the nearest neighbour of $p_{i}^{q}$ from the key view, which can enhance the structured neighbourhood information.
Specifically, for each vectors in $p_{i}^{q}$, its associated nearest neighbour can be obtained by applying an \texttt{argmax} operation to the similarity of $z^{q}$ and $z^{k}$,
which is formulated as:
\begin{equation}
    n_{i} = \mathop{\arg\max}\limits_{j} sim(z_{i}^{q},z_{j}^{k}),\, z_{i}^{q} \in z^{q},\, z_{j}^{k} \in z^{k},
    \label{eq5}
\end{equation}
where $sim(u,v) = u^\top v / \Vert u \Vert \Vert v \Vert$ denotes the cosine similarity,
$i$ is the spatial index of attentional query features $p_{i}^{q}$, $j$ is the spatial index of key features $z_{j}^{k}$,
and $n_{i}$ means the obtained nearest neighbour of $p_{i}^{q}$.

Finally, we get the evolved corresponding set $c_{i}$ for each $p_{i}^{q}$, which is formulated as:
\begin{equation}
    c_{i} = s_{i} \cup n_{i}.
    \label{eq7}
\end{equation}

\myparagraph{Similarity Learning Objectives.}
As illustrated in Figure~\ref{fig:method}, given attentional query vectors $p_{i}^{q}$ and the corresponding set $c_{i}$,
the positive pairs can be directly obtained and negatives $k_{-}$ is provided by a queue of global average-pooled feature in the key view.
The set-level contrastive loss is calculated as:
\begin{equation}
\resizebox{.9\hsize}{!}{
    $\mathcal{L}_{set} = \sum_{i} \frac{-1}{|c_{i}|} \sum_{j \in c_{i}} \log \frac{\exp(p_{i}^{q} \cdot p_{j}^{k}/\tau)}{\sum_{j\in c_{i}} \exp(p_{i}^{q} \cdot p_{i}^{k}/\tau) + \sum_{k_{-}} \exp(p_{i}^{q} \cdot k_{-}/\tau)}$,
}
\label{eq8}
\end{equation}
where the $\cdot$ symbol denotes the inner product, $|c_{i}|$ is a cardinality of $c_{i}$, and $p_{j}^{k}$ is the attentional vector in the key view.
$\tau$ is a temperature hyper-parameter.
Following \cite{chen2020improved}, we set $\tau=0.2$ as default.
Overall, the total loss in our framework can be formulated as follow,
\begin{equation}
    \mathcal{L} = (1 - \lambda)\mathcal{L}_{img} + \lambda\mathcal{L}_{set}.
    \label{eq9}
\end{equation}
where $\lambda$ is a hyper-parameter to balance two terms, which is set to 0.5 \cite{wang2020DenseCL}.


\section{Experiments}\label{sec:Experiments}
\begin{figure*}[t]
    \begin{minipage}[t]{0.48\textwidth}
    \begin{center}
    \makeatletter\def\@captype{table}\makeatother\caption{\textbf{Building SetSim on various self-supervised learning frameworks.} 
     All methods are pretrained for 200 epochs on the IN-100 dataset and fine-tuned on PASCAL VOC object detection.
     The significant improvements indicate that our method is applicable to multiple frameworks. (Average over 5 trials)}
    \setlength{\tabcolsep}{2mm}{
    \begin{tabular}{l|lll}
        \toprule[1.2pt]
        method & AP$^{b}$ & AP$^{b}_{50}$ & AP$^{b}_{75}$ \\
        \midrule[1.2pt]
        MoCo-v2         & 54.3 & 80.3 & 60.2 \\
        + SetSim & \textbf{56.1} \textcolor{green}{(+1.8)} & \textbf{81.6} \textcolor{green}{(+1.3)} & \textbf{62.5} \textcolor{green}{(+2.3)} \\
        \midrule
        SimSiam         & 54.5 & 80.4 & 60.5 \\
        + SetSim & \textbf{55.8} \textcolor{green}{(+1.3)} & \textbf{81.2} \textcolor{green}{(+0.8)} & \textbf{62.0} \textcolor{green}{(+1.5)} \\
        \bottomrule[1.2pt]
    \end{tabular}}
    \label{tab:baseline}
    \end{center}
    \end{minipage}
    \quad
    \begin{minipage}[t]{0.48\textwidth}
        \begin{center}
        \makeatletter\def\@captype{table}\makeatother\caption{\textbf{Ablation study of matching strategy.}
        All methods are pretrained for 200 epochs on the IN-100 dataset and fine-tuned on PASCAL VOC object detection.
        Note that the first three are pixel-to-pixel matching strategies, so the number of selected pixels across two views needs to remain the same.
        (Average over 5 trials)}
        \setlength{\tabcolsep}{4mm}{
        \begin{tabular}{l|ccc}
            \toprule[1.2pt]
            strategy & AP$^{b}$ & AP$^{b}_{50}$ & AP$^{b}_{75}$ \\
            \midrule[1.2pt]
            $Random$    & 54.9 & 80.6 & 60.6 \\
            $Sort$      & 55.4 & 81.0 & 61.1 \\
            $Hungarian$ & 55.6 & 81.1 & 61.5 \\
            \midrule
            $Set2Set$   & 56.0 & 81.4 & 61.9 \\
            $Set2Set-NN$   & \textbf{56.1} & \textbf{81.6} & \textbf{62.0} \\
            \bottomrule[1.2pt]
        \end{tabular}
        \label{tab:ablation_strategy}}
        \end{center}
    \end{minipage}
\end{figure*} 
\begin{figure*}[t]
    \begin{minipage}[h]{0.48\textwidth}
        \begin{center}
            \includegraphics[scale=0.44]{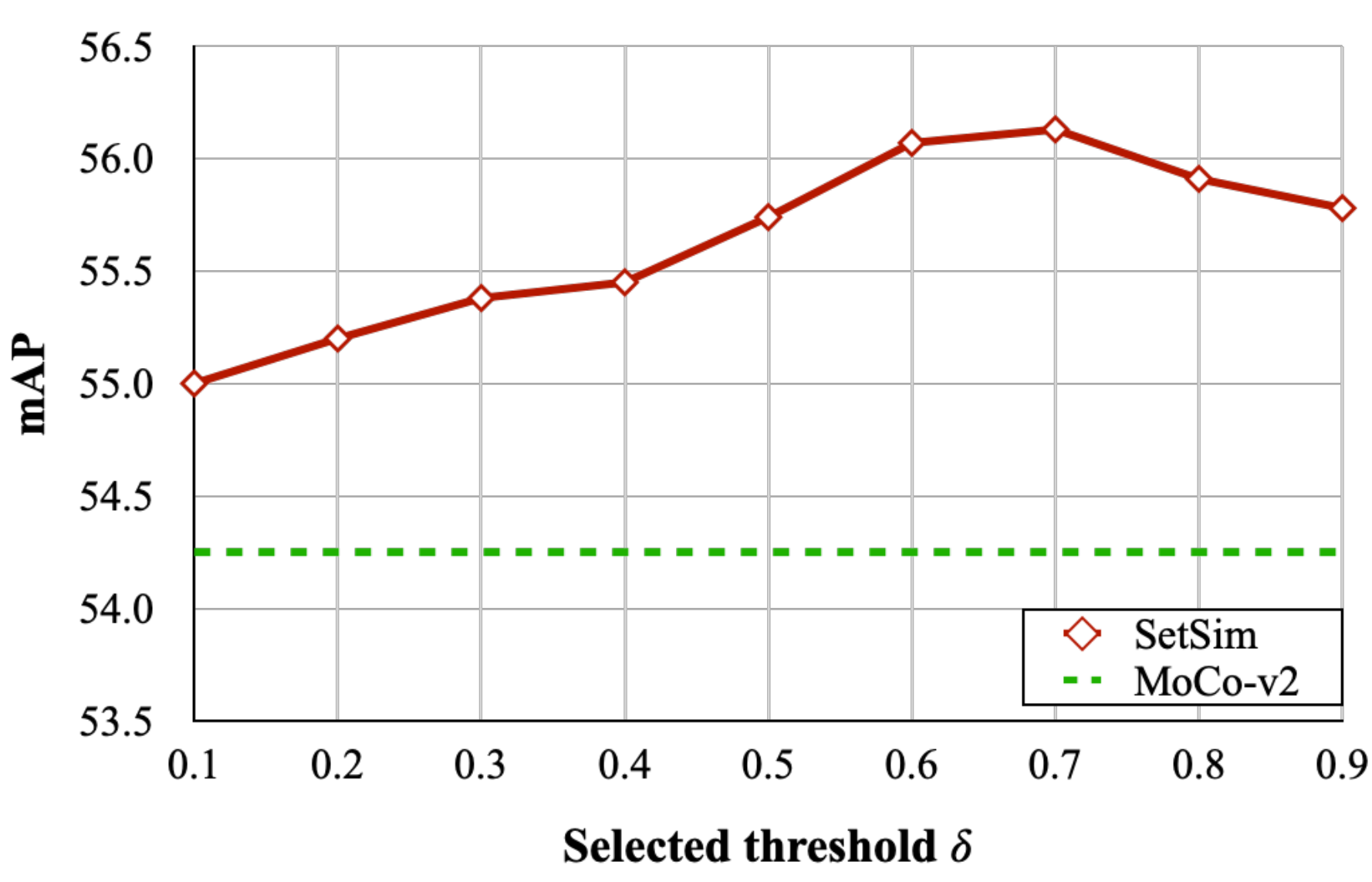}
            \setlength{\abovecaptionskip}{0.7pt}
            \makeatletter\def\@captype{figure}\makeatother\caption{\textbf{Ablation study of selected threshold.}
            Each model is pretrained on the IN-100 dataset for 200 epochs and fine-tuned on PASCAL VOC object detection.
            (Average over 5 trials)}
        \label{fig:ablation_ratio} 
        \end{center}
    \end{minipage}
    \quad
    \begin{minipage}[h]{0.48\textwidth}
    \begin{center}
    \makeatletter\def\@captype{table}\makeatother\caption{\textbf{Ablation study of the combination of correspondences.} 
    ``$Set$'' and ``$Geo$'' denote the set-based and geometry-based correspondences, respectively. 
    ``$Sym$'' denotes the symmetrized loss \cite{chen2020exploring}.
    Each model is pretrained on the IN-100 dataset for 200 epochs and fine-tuned on PASCAL VOC object detection.
    The baseline is MoCo-v2.}
    \setlength{\tabcolsep}{5mm}{
    \begin{tabular}{ccc|l}
        \toprule[1.2pt]
        $Set$ & $Geo$ & $Sym$ & AP$^{b}$ \\
        \midrule[1.2pt]
         &  &  & 54.3 \\
        \ding{51} &  &  & 55.3 \textcolor{green}{(+1.0)}\\
         & \ding{51} &  & 55.2 \textcolor{green}{(+0.9)}\\
        \ding{51} & \ding{51} &  & 56.0 \textcolor{green}{(+1.7)}\\
        \ding{51} & \ding{51} & \ding{51} & \textbf{56.1} \textcolor{green}{(+1.8)}\\
        \bottomrule[1.2pt]
    \end{tabular}
    \label{tab:ablation_correspondence}}
    \end{center}
    \end{minipage}
\end{figure*} 

We conduct the experiments of self-supervised pretraining on two type of ImageNet dataset \cite{deng2009imagenet}:
(a) IN-1K contains \textasciitilde1.25M images,
(b) IN-100 \cite{tian2020contrastive} is a subset of ImageNet-1K containing \textasciitilde125K images.
Subsequently, we evaluate the transfer performance on various dense prediction tasks.
In particular, the pretrained model is fine-tuned on PASCAL VOC \cite{everingham2010pascal} for object detection and semantic segmentation,
COCO \cite{lin2014microsoft} for object detection, instance segmentation and keypoint detection,
Cityscapes \cite{cordts2016cityscapes} for semantic segmentation and instance segmentation,
and ADE20K \cite{zhou2017scene} for semantic segmentation.
Considering the efficiency, the ablation study is conducted on the IN-100 dataset.
Following the common protocol \cite{he2020momentum,chen2020improved}, we report the 200-epoch pretrained model to compare with state-of-the-art methods.
Note that all the comparing 200-epoch pretrained weights are downloaded from their official releases respectively except for PixPro,
which didn't offer the 200-epoch pretrained weights, is re-trained by using their official code.

\myparagraph{Pretraining Setting.}
Following the setting in \cite{chen2020exploring,he2020momentum,chen2020improved},
we utilize SGD as our optimizer with initial learning rates of 0.03 for MoCo-v2 and 0.1 for SimSiam.
The learning rate is updated under the cosine decay scheduler \cite{loshchilov2016sgdr,chen2020simple}.
The weight decay and momentum in SGD are set as 0.0001 and 0.9, respectively. 
Shuffling BN is adopted for MoCo, and synchronized BN is adopted for SimSiam during pre-training.
Each pre-trained model is trained on 8 Tesla V100 GPUs with the batch size of 256. 
Compared with the baselines, our method slightly increases the  training time. 
Note that our method does not cause any extra computational cost on following downstream tasks.

\subsection{Ablation Study}

\begin{figure*}[t]
    \begin{minipage}[t]{0.46\textwidth}
     \begin{center}
         \makeatletter\def\@captype{table}\makeatother\caption{\textbf{Comparisons with the state-of-the-art approaches on PASCAL VOC object detection.}
         A Faster R-CNN (R50-C4) \cite{ren2015faster,wu2019detectron2} is trained on \texttt{trainval07+12}, evaluated on \texttt{test07}.
         Each model is pretrained for 200 epochs.
         (Average over 5 trials)}
         \setlength{\belowcaptionskip}{0.1mm}
         \setlength{\tabcolsep}{3.5mm}{
     \begin{tabular}{l|c|cc}
         \toprule[1.2pt]
         Method & AP$^{b}$ & AP$^{b}_{50}$ & AP$^{b}_{75}$ \\
         \midrule[1.2pt]
         \textcolor{gray}{Random init.}         & \textcolor{gray}{33.8} & \textcolor{gray}{60.2} & \textcolor{gray}{33.1} \\
         IN-1K sup.                             & 53.5 & 81.3 & 58.8 \\
         \midrule
         SimCLR \cite{chen2020simple}          & 51.5 & 79.4 & 55.6 \\
         MoCo-v2 \cite{chen2020improved}       & 57.0 & 82.3 & 63.3 \\
         BYOL \cite{grill2020bootstrap}        & 55.3 & 81.4 & 61.1 \\
         SimSiam \cite{chen2020exploring}      & 56.4 & 82.0 & 62.8 \\
         SwAV \cite{caron2020unsupervised}     & 55.4 & 81.5 & 61.4 \\
         InfoMin \cite{tian2020makes}          & 57.5 & 82.5 & 64.0 \\
         DenseCL \cite{wang2020DenseCL}        & 58.7 & 82.8 & 65.2 \\
         PixPro \cite{xie2020propagate}        & \textbf{59.4} & 83.1 & \textbf{66.9} \\
         ReSim-C4 \cite{xiao2021region}        & 58.7 & 83.1 & 66.3 \\
         \midrule
         SetSim                                & 59.1 & \textbf{83.2} & 66.1 \\
         \bottomrule[1.2pt]
     \end{tabular}
     \label{tab:voc_det}}
     \end{center}
     \end{minipage}
     \quad
     \begin{minipage}[t]{0.48\textwidth}
     \begin{center}
     \makeatletter\def\@captype{table}\makeatother\caption{\textbf{Comparisons with the state-of-the-art approaches on Cityscapes, PASCAL VOC, and ADE20K semantic segmentation.}
     A FCN (R50) \cite{long2015fully,mmseg2020} is adopted for all methods.
     Following the setting of DenseCL \cite{wang2020DenseCL}, we fine-tune all methods with their official pretrained weights.
     Each model is pretrained for 200 epochs.
     (Average over 5 trials)}
     \setlength{\belowcaptionskip}{-0.3mm}
     \setlength{\tabcolsep}{3.5mm}{
     \begin{tabular}{l|c|c|c}
         \toprule[1.2pt]
         \multirow{2}{*}{Method} & \multicolumn{3}{c}{mIoU} \\\noalign{\smallskip}
         \cline{2-4}\noalign{\smallskip}
         ~ & Citys & VOC & ADE \\
         \midrule[1.2pt]
         \textcolor{gray}{Random init.}         & \textcolor{gray}{65.1} & \textcolor{gray}{40.7} & \textcolor{gray}{29.4} \\
         IN-1K sup.                             & 74.0 & 67.5 & 35.9 \\
         \midrule
         MoCo-v1 \cite{he2020momentum}         & 74.5 & 66.2 & 37.0 \\
         MoCo-v2 \cite{chen2020improved}       & 75.4 & 67.3 & 36.9\\
         SwAV \cite{caron2020unsupervised}     & 73.2 & 65.2 & 36.7 \\
         DenseCL \cite{wang2020DenseCL}        & 75.9 & 68.9 & 38.1 \\
         PixPro \cite{xie2020propagate}        & 76.0 & 70.3 & 38.3 \\
         ReSim-C4 \cite{xiao2021region}        & 75.8 & 68.4 & 37.9 \\
         \midrule
         SetSim                                & \textbf{77.0} & \textbf{70.9} & \textbf{38.6} \\
         \bottomrule[1.2pt]
     \end{tabular}
     \label{tab:citys_seg}}
     \end{center}
     \end{minipage}
 \end{figure*}

We first build our method on various self-supervised learning baseline to evaluate the effectiveness and expansibility. 
To study each component in SetSim, we conduct extensive experiments with different setting and visualize the learned set-wsie correspondences for qualitative analysis. 
Due to the extra training time caused by SimSiam \cite{chen2020exploring}, we finally adopt the MoCo-v2 \cite{chen2020improved} as our base framework to compare with the state-of-the-art methods on various dense prediction tasks. 

\myparagraph{Experimental Setting.}
In this section, we fine-tune each pre-trained model on the widely-used PASCAL VOC object detection \cite{everingham2010pascal}. 
Following \cite{he2020momentum},
we train a Fast R-CNN detector (C4-backbone) \cite{ren2015faster,wu2019detectron2} on \texttt{trainval07+12} set (\textasciitilde16.5k images) with standard 24k iterations,
and evaluate on \texttt{test2007} set (\textasciitilde4.9k images). 
During the training process, the short-side length of input images is randomly selected from 480 to 800 pixels and fixed at 800 for inference.
Similar to \cite{he2020momentum,chen2020improved}, we fine-tune all Batch Normalization layers and adopt synchronized version during training. 
All results are averaged over five trials to overcome the randomness.

\myparagraph{Comparisons with Baselines.}
Firstly, we compare our method with the MoCo-v2 baseline. Note that the overall architecture is similar to MoCo-v2, ensuring a fair comparison. 
As shown in Table~\ref{tab:baseline}, SetSim can effectively explore spatial representations across two views' convolutional features.
The most significant performance gap occurs at AP$^{b}_{75}$, a high AP metric, which demonstrates that SetSim can improve the localization capability. 
Besides, we also build SetSim on a recently proposed self-supervised learning framework, SimSiam, without considering negative samples. 
Following \cite{chen2020exploring}, we keep SimSiam's design of the projector and predictor for image-level representation learning. 
And, the convolutional projector and predictor are parallel with the image-level part for dense representation learning. 
As illustrated in Table~\ref{tab:baseline}, SetSim surpasses the MoCO-v2 baseline by a large margin of 1.8, 1.3, 2.3 at AP$^{b}$, AP$^{b}_{50}$, and AP$^{b}_{75}$, which powerfully demonstrates the effectiveness and expansibility of our method. 
Meanwhile, we observe the similar phenomenon in the SimSiam experiment. 

\myparagraph{Matching Strategy.} 
As illustrated in Table~\ref{tab:ablation_strategy}, we compare five different matching strategies.
The first three are pixel-to-pixel matching strategies.
(1) $Random$: features from two views are randomly matched.
(2) $Sort$: two views' features are sorted from large to small based on their attention values, then matched one by one.
(3) $Hungarian$: features from two views are matched by the Hungarian algorithm \cite{kuhn1955hungarian}, where we set the cosine distance as the cost matrix. 
$Random$ can obtain 0.6\% AP$^{b}$ gains compared to MoCo-v2, which can be explained as the convolutional projector is able to keep the spatial information and some random correspondences are correct.
$Sort$ and $Hungarian$ bring more improvements than the former, demonstrating that these two strategies can effectively establish more adequate pixel-wise correspondences across two views.
$Set2Set$ performs the better AP$^{b}$ of 56.0\% than three pixel-to-pixel strategies,
because $Set2Set$ is able to learn more spatial-structured information by pulling closer two corresponding sets of pixels across views.
Compared to pixel-wise similarity learning,  sets-level representation learning facilitates exploring robust visual representation because the set contains more semantic and structure information.
Finally, $Set2Set-NN$ can further improve AP$^{b}$ and AP$^{b}_{50}$ by 0.1\% and 0.2\%, because $NN$ can recycle useful features excluded from sets and enhances the spatial-structured information.

\myparagraph{Selected threshold $\delta$.}
We explicitly control the selected threshold $\delta$ of attentional vectors to study the effects.
Specifically, we raise $\delta$ from 0.1 to 0.9 to limit the number of selected attentional feature during pixel-level representation learning. 
As shown in Figure~\ref{fig:ablation_ratio}, we can observe a trend that increasing $\delta$ from 0.1 to 0.7 brings significant gains of 1.13\% AP$^{b}$.
It is explainable that the lower $\delta$ leads to the participation of more attentional vectors, which are not related to salient objects, in the corresponding set, thus establishing incorrect spatial correspondences.
As the training progresses, the model gradually memorizes these incorrect correspondences, which causes the limitation of transfer ability on downstream tasks.
Besides, the performance slightly drops 0.35\% AP$^{b}$ if we further increase $\delta$ to 0.9, because too few attentional vectors are adopted to learn sufficient visual representations.

\myparagraph{Combination of Spatial Correspondence.}
As shown in Table~\ref{tab:ablation_correspondence}, we further investigate the effect of different type of spatial correspondence.
It can be seen that both ``Set'' and ``Geo'' \cite{xie2020propagate} can achieve better transfer performance than MoCo-v2 (55.3 and 55.2 \textit{v.s.} 54.3). 
When two types of spatial correspondence are adopted simultaneously, the model can obtain significant gains of 1.8\% than the baseline,
which indicates that the two types of spatial correspondence are complementary.
Specifically, ``Set'' exploits sets' semantic and spatial-structured information, and ``Geo'' can force geometry-corresponding local representations to remain constant over different viewing conditions.
Finally, we adopt the symmetrized loss \cite{chen2020exploring} to improve performance, and choose this as the default for SetSim.

\begin{table*}[t]
    \setlength{\belowcaptionskip}{2.7pt}
    \begin{center}
    \caption{\textbf{Comparisons with the state-of-the-art approaches on COCO object detection, instance segmentation, and keypoint detection.} 
    All methods are fine-tuned on \texttt{train2017} with 1$\times$ schedules and evaluated on \texttt{val2017}. 
    A Mask-RCNN (R50) \cite{he2017mask,wu2019detectron2} with FPN \cite{lin2017feature} is adopted for all methods. 
    Average precision on bounding-boxes (AP$^{b}$), masks (AP$^{m}$) and keypoint (AP$^{kp}$) are used as benchmark metrics. 
    Following \cite{he2020momentum}, we fine-tune DenseCL and PixPro with their official pretrained weights, because they adopted a different COCO fine-tuning setting from the common approach \cite{he2020momentum,chen2020improved,chen2020simple}.
    Each model is pretrained for 200 epochs.
    (Average over 5 trials)}
    \setlength{\tabcolsep}{3.7mm}{
    \begin{tabular}{l|ccc|ccc|ccc}
        \toprule[1.2pt]
        \multirow{2}{*}{Method} & \multicolumn{3}{c|}{Object Det.} & \multicolumn{3}{c|}{Instance Seg.} & \multicolumn{3}{c}{Keypoint Det.} \\
        \cline{2-10}
        ~ & AP$^{b}$ & AP$^{b}_{50}$ & AP$^{b}_{75}$ & AP$^{m}$ & AP$^{m}_{50}$ & AP$^{m}_{75}$ & AP$^{kp}$ & AP$^{kp}_{50}$ & AP$^{kp}_{75}$ \\
        \midrule[1.2pt]
        \textcolor{gray}{Random init.}         & \textcolor{gray}{31.0} & \textcolor{gray}{49.5} & \textcolor{gray}{33.2} & \textcolor{gray}{28.5} & \textcolor{gray}{46.8} & \textcolor{gray}{30.4} & \textcolor{gray}{63.0} & \textcolor{gray}{85.1} & \textcolor{gray}{68.4} \\
        IN-1K sup.                             & 38.9 & 59.6 & 42.7 & 35.4 & 56.5 & 38.1 & 65.3 & 87.0 & 71.3 \\
        \midrule
        MoCo-v1 \cite{he2020momentum}         & 38.5 & 58.9 & 42.0 & 35.1 & 55.9 & 37.7 & 66.1 & 86.7 & 72.4 \\
        MoCo-v2 \cite{chen2020improved}       & 38.9 & 59.2 & 42.4 & 35.4 & 56.2 & 37.8 & 66.3 & 87.1 & 72.2 \\
        VADeR \cite{pinheiro2020unsupervised} & 39.2 & 59.7 & 42.7 & 35.6 & 56.7 & 38.2 & 66.1 & 87.3 & 72.1 \\
        DenseCL \cite{wang2020DenseCL}        & 39.4 & 59.9 & 42.7 & 35.6 & 56.7 & 38.2 & 66.6 & 87.4 & \textbf{72.6} \\
        PixPro \cite{xie2020propagate}        & 39.8 & 59.5 & 43.7 & 36.1 & 56.5 & 38.9 & 66.5 & 87.6 & 72.3 \\
        ReSim-C4 \cite{xiao2021region}        & 39.3 & 59.7 & 43.1 & 35.7 & 56.7 & 38.1 & 66.3 & 87.2 & 72.4 \\
        \midrule
        SetSim                                 & \textbf{40.2} & \textbf{60.7} & \textbf{43.9} & \textbf{36.4} & \textbf{57.7} & \textbf{39.0} & \textbf{66.7} & \textbf{87.8} & 72.4 \\
        \bottomrule[1.2pt]
    \end{tabular}
    \label{tab:coco_det_ins}}
    \end{center}
\end{table*}

\begin{figure*}[t]
    \begin{center}
        \includegraphics[scale=0.44]{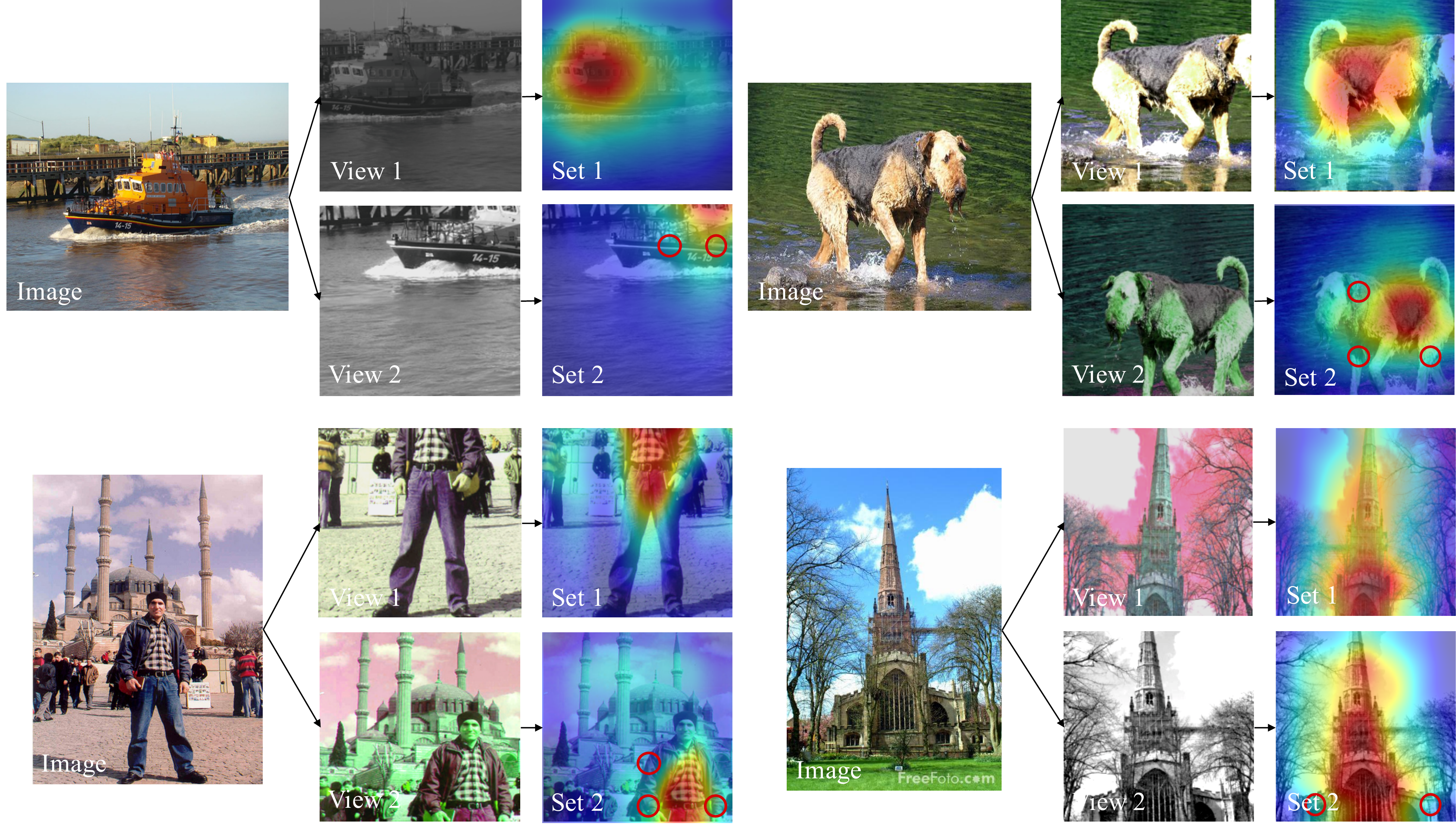}
        \setlength{\abovecaptionskip}{0.7pt}
        \caption{\textbf{Visualization of corresponding sets across two data-augmented views.}
        Each corresponding set is constructed by 200-epoch pretrained model.
        \textcolor{red}{Red circle} in Set 2 denotes the Set 1's nearest neighbours in the View 2.}
    \label{fig:attention} 
    \end{center}
    \vspace{-2.0mm}
\end{figure*}

\myparagraph{Qualitative Analysis.}
As illustrated in Figure~\ref{fig:attention},
we visualize the learned corresponding sets between two views of input images.
The visualization shows that SetSim can accurately focus on salient objects or regions with similar semantic information across two data-augmented views, which keeps the coherence of input images.
Besides, SetSim effectively filters out misleading local features, which avoids establishing noisy spatial correspondence.
From a qualitative perspective, the visualization confirms our motivation for exploring set similarity across two views to improve the robustness.

\subsection{Main Results}

\myparagraph{PASCAL VOC Object Detection.}
We utilize the Faster R-CNN (R50-C4) detector and keep the same setting as mentioned in Section 4.1.
As illustrated in Table~\ref{tab:voc_det}, we report the object detection result on PASCAL VOC and compare it with a series of state-of-the-art methods.
Our method yields significant improvements than the MoCo-v2 baseline \cite{he2020momentum} at AP$^{b}$, AP$^{b}_{50}$, and AP$^{b}_{75}$.
Furthermore, our method surpasses DenseCL \cite{wang2020DenseCL} and ReSim \cite{xiao2021region} by 0.4\% at AP$^{b}$, respectively.

\myparagraph{PASCAL VOC \& Cityscapes \& ADE20K Semantic Segmentation.}
Following the setting of \cite{wang2020DenseCL}\footnote{\url{https://github.com/WXinlong/mmsegmentation}}, we fine-tune an FCN \cite{long2015fully,mmseg2020} on VOC \texttt{train\_aug2012} set (\textasciitilde10k images) for 20k iterations and evaluate on \texttt{val2012} set.
Besides, We fine-tune on Cityscapes \texttt{train\_fine} set (2975 images) for 40k iterations and test on \texttt{val} set.
Finally, following the standard scheduler \cite{mmseg2020}, we fine-tune an FCN on ADE20K \cite{zhou2017scene} \texttt{train} set (\textasciitilde20k images) for 80k iterations and evaluate on \texttt{val} set (\textasciitilde2k images).
As shown in Table~\ref{tab:citys_seg}, SetSim obtains remarkable gains of 3.0\%, 3.4\%, and 2.7\% mIoU on Cityscapes, VOC, and ADE20K than the supervised ImageNet pre-training, respectively.
Moreover, SetSim outperforms DenseCL \cite{wang2020DenseCL} and ReSim \cite{xiao2021region} by significant margins on three benchmarks, which strongly demonstrate our method is friendly for dense prediction task.

\begin{table}[t]
    \begin{center}
        \caption{\textbf{Results on Cityscapes instance segmentation.}
        A Mask-RCNN (R50) \cite{he2017mask,wu2019detectron2} with FPN \cite{lin2017feature} is fine-tuned on \texttt{train\_fine} set and evaluated on \texttt{val} set.
        Following \cite{he2020momentum}, we fine-tune all methods with their official pretrained weights because the results of SwAV, DenseCL, and PixPro are not public.
        Each model is pretrained for 200 epochs.
        (Average over 5 trials)}
    \setlength{\tabcolsep}{6mm}{
        \begin{tabular}{l|cc}
            \toprule[1.2pt]
            Method & AP$^{m}$ & AP$^{m}_{50}$ \\
            \midrule[1.2pt]
            \textcolor{gray}{Random init.}         & \textcolor{gray}{25.6} & \textcolor{gray}{51.5} \\
            IN-1K sup.                             & 32.9 & 59.6 \\
            \midrule
            MoCo-v1 \cite{he2020momentum}         & 32.8 & 59.2 \\
            MoCo-v2 \cite{chen2020improved}       & 33.4 & 60.4 \\ 
            SwAV \cite{caron2020unsupervised}     & 33.6 & 62.5 \\
            DenseCL \cite{wang2020DenseCL}        & 34.9 & 62.5 \\ 
            PixPro \cite{xie2020propagate}        & 34.0 & 62.0 \\
            ReSim-C4 \cite{xiao2021region}        & 35.4 & 63.1 \\
            \midrule
            SetSim                                & \textbf{35.6} & \textbf{63.4} \\
            \bottomrule[1.2pt]
    \end{tabular}
    \label{tab:citys_ins}}
    \end{center}
\end{table}

\myparagraph{COCO Object Detection \& Instance Segmentation \& Keypoint Detection.}
Following \cite{he2020momentum}, we fine-tune a Mask-RCNN (R50) \cite{he2017mask,wu2019detectron2} with FPN \cite{lin2017feature} (Keypoint-RCNN \cite{wu2019detectron2}) under $1\times$ scheduler and add new Batch Normalization layers before the FPN parameters.
All Batch Normalization statistics are synchronized across GPUs.
Training is conducted on \texttt{train2017} split with \textasciitilde118k images, and testing is conducted on \texttt{val2017} split.
The short-side length of input images is randomly selected from 640 to 800 pixels during training and fixed at 800 pixels for testing.
Following \cite{he2020momentum,chen2020improved,chen2020exploring}, Average Precision on bounding-boxes (AP$^{b}$) and Average Precision on masks (AP$^{m}$) are adopted as our metrics.
Table~\ref{tab:coco_det_ins} demonstrates that SetSim improve over the state-of-the-art PixPro \cite{xie2020propagate} under both box and mask metrics, where the gains are 0.4\%, 0.3\%, and 0.2\% at AP$^{b}$, AP$^{m}$, and AP$^{kp}$. 

\myparagraph{Cityscapes Instance Segmentation.}
Following the setting of ReSim \cite{xiao2021region},
we fine-tune a Mask-RCNN (R50) \cite{he2017mask,wu2019detectron2} with FPN \cite{lin2017feature} and add Batch Normalization layers before the FPN,
and synchronize all Batch Normalization during training.
Table~\ref{tab:citys_ins} illustrates the comparisons of SetSim versus the supervised pre-training counterparts and a series of state-of-the-art methods.
SetSim remarkably improves over the baseline MoCo-v2 by 2.2\% and 3.0\% at AP$^{m}$ and AP$^{m}_{50}$ and surpass the state-of-the-art work, ReSim \cite{xiao2021region}. We present lots of qualitative analysis in the supplementary.


\section{Conclusion}

In this paper, we have proposed a simple but effective dense self-supervised representation learning framework, SetSim, by exploring set similarity across views for dense prediction tasks.
By resorting to attentional features, SetSim constructs the corresponding set across two views, which alleviates the effect of misleading pixels and semantic inconsistency.
Besides, since some useful features are neglected from sets, we further search the cross-view nearest neighbours of sets to enhance the structure neighbour information.
Finally, a contrastive/similarity loss function is utilized to map two sets of pixel-wise feature vectors to similar representations in the embedding space, encouraging the model to learn adequate dense visual representations.
Compared to the MoCo-v2 and SimSiam baseline, our SetSim significantly improves the localization and classification abilities on a series of dense prediction tasks, which narrow the transfer gap between the self-supervised pretraining and dense prediction tasks.
Empirical evaluations have demonstrated that SetSim is comparable with or outperforms state-of-the-art approaches on various downstream tasks, including PASCAL VOC object detection, COCO object detection, COCO instance segmentation, COCO keypoint detection, Cityscapes instance segmentation, PASCAL VOC semantic segmentation, Cityscapes semantic segmentation, and ADE20K semantic segmentation.
In the future, we will further investigate how to effectively mine semantic information from multi-view inputs in a self-supervised manner.

\textbf{Acknowledgements}\quad We thank Wanjing Zong and Ziyu Chen for the helpful discussions on this work. Meanwhile, We appreciate Weiqiong Chen, Bin Long and Rui Sun for AWS technical support.


{\small
  \bibliographystyle{ieee_fullname}
  \bibliography{sections/mainbib}
}



\end{document}